\documentclass[final]{cvpr}

\usepackage{times}
\usepackage{epsfig}
\usepackage{graphicx}
\usepackage{amsmath}
\usepackage{amssymb}
\usepackage{enumitem}
\usepackage{float}
\usepackage{bm}
\usepackage{booktabs}
\usepackage{multirow}

\newcommand\blfootnote[1]{%
  \begingroup
  \renewcommand\thefootnote{}\footnote{#1}%
  \addtocounter{footnote}{-1}%
  \endgroup
}

\usepackage[pagebackref=true,breaklinks=true,colorlinks,bookmarks=false]{hyperref}

\def\cvprPaperID{3444} 
\def\confYear{CVPR 2021}

\graphicspath{{figures/}}

\newcommand{\myparagraph}[1]{\textbf{#1}}

\begin{document}

\title{\bf \Large CDFI: Compression-Driven Network Design for Frame Interpolation\vspace{.2in}}

\author{Tianyu Ding$^{1*}$
\and
Luming Liang$^{2*\dagger}$
\and
Zhihui Zhu$^3$
\and
Ilya Zharkov$^2$
}

\date{$^1$Johns Hopkins University \qquad  $^2$Microsoft \qquad $^3$University of Denver\\
{\tt\small tding1@jhu.edu},
{\tt\small \{lulian,zharkov\}@microsoft.com},
{\tt\small zhihui.zhu@du.edu}
}

\maketitle

\begin{abstract}

DNN-based frame interpolation---that generates the intermediate frames given two consecutive frames---typically relies on heavy model architectures with a huge number of features, preventing them from being deployed on systems with limited resources, \eg, mobile devices. 
We propose a compression-driven network design for frame interpolation (CDFI), that leverages model pruning through sparsity-inducing optimization to significantly reduce the model size while achieving superior performance. Concretely, we first compress the recently proposed AdaCoF model and show that a 10$\times$ compressed AdaCoF performs similarly as its original counterpart; then we further improve this compressed model by introducing a multi-resolution warping module, which boosts visual consistencies with multi-level details. As a consequence, we achieve a significant performance gain with only a quarter in size compared with the original AdaCoF. Moreover, our model performs favorably against other state-of-the-arts in a broad range of datasets. Finally, the proposed compression-driven framework is generic and can be easily transferred to other DNN-based frame interpolation algorithm. Our source code is available at \url{https://github.com/tding1/CDFI}.

\end{abstract}

\section{Introduction}

\blfootnote{$^*$Equal contribution. This work was done when Tianyu Ding was an intern at Applied Sciences Group, Microsoft.}
\blfootnote{$^\dagger$Corresponding author.}\hspace{-.06in}Video frame interpolation is a lower level computer vision task referring to the generation of intermediate (non-existent) frames between actual frames in a sequence, which is able to largely increase the temporal resolution. It plays an important role in many applications, including frame rate up-conversion~\cite{bao2018high}, slow-motion generation~\cite{jiang2018super}, and novel view synthesis~\cite{flynn2016deepstereo,zhou2016view}. Though fundamental, the problem is challenging in that the complex motion, occlusion and feature variation in real world videos are difficult to estimate and predict in a transparent way.

\begin{figure}[]
    \centering
    \includegraphics[width=0.47 \textwidth]{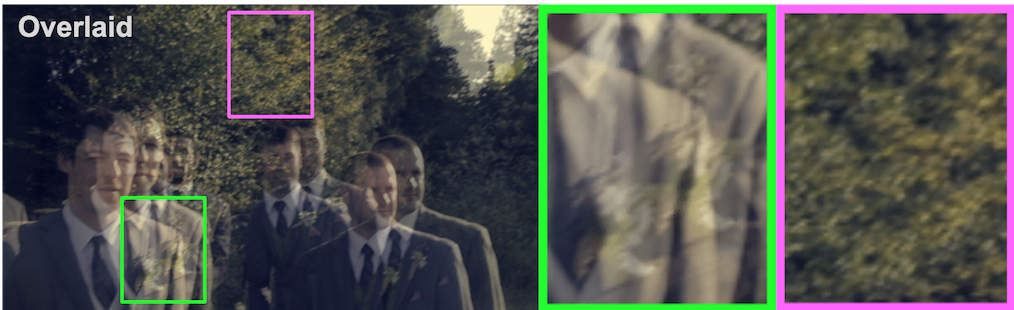}\\
    \includegraphics[width=0.47 \textwidth]{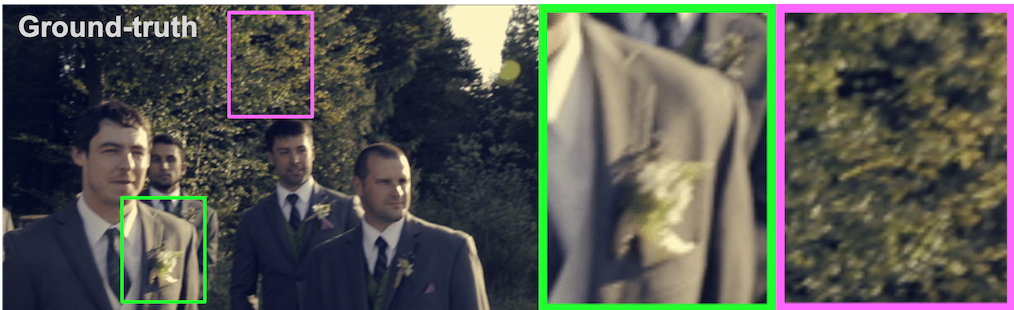}\\
     \includegraphics[width=0.47 \textwidth]{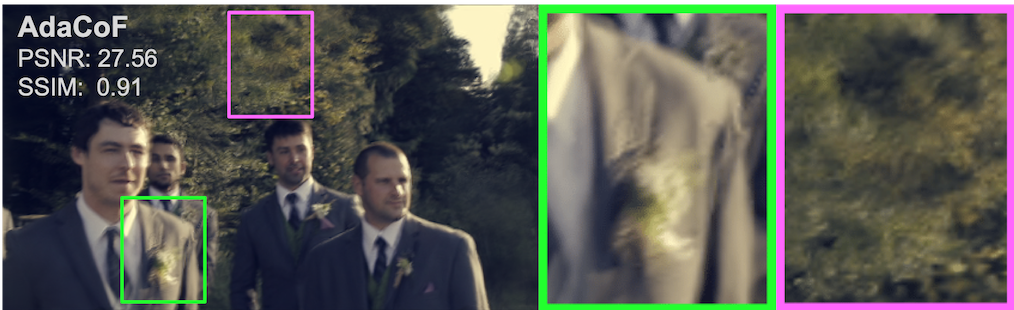}\\
     \includegraphics[width=0.47 \textwidth]{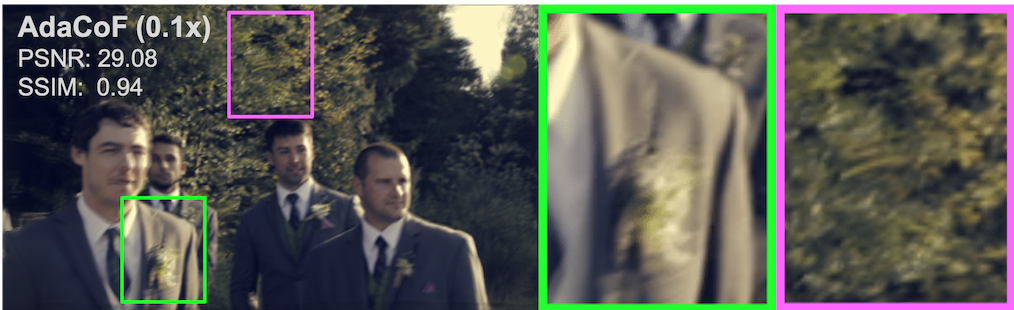}\\
     \includegraphics[width=0.47 \textwidth]{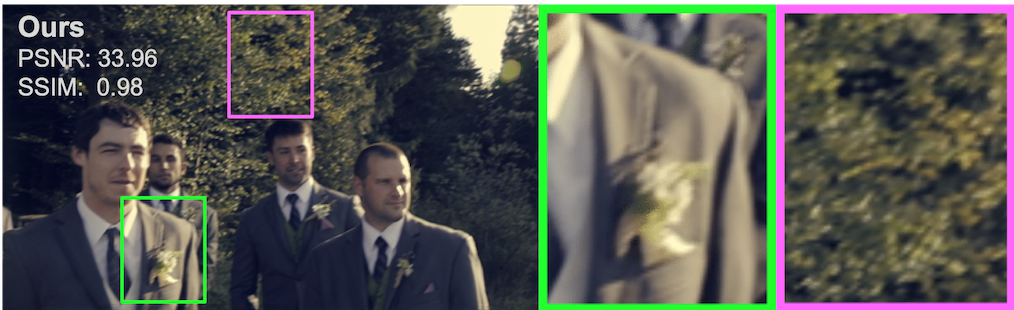}
    \caption{\textbf{A challenging example consists of large motion, severe occlusion and non-stationary finer details.} From top to bottom: the overlaid two inputs, the ground-truth middle frame, the frame generated by AdaCoF~\cite{lee2020adacof},
    the frame generated by the 10$\times$ compressed AdaCoF, and the frame generated by our method. The compressed AdaCoF even outperforms the full one in this case.}
    \label{fig:example}
    \vspace{-.25in}
\end{figure}

Recently, a large number of researches have been conducted in this area, especially those based on deep neural networks (DNN) for their promising results in motion estimation~\cite{dosovitskiy2015flownet,ilg2017flownet,sun2018pwc,weinzaepfel2013deepflow}, occlusion reasoning~\cite{bao2019depth,jiang2018super,peleg2019net} and image synthesis~\cite{dosovitskiy2015learning,flynn2016deepstereo, kalantari2016learning,kulkarni2015deep,zhou2016view}. In particular, due to the rapid expansion in optical flow~\cite{baker2011database, werlberger2011optical}, many approaches either utilize an off-the-shelf flow model~\cite{bao2019depth,niklaus2018context,niklaus2020softmax,xu2019quadratic} or estimate their own task-specific flow~\cite{jiang2018super,liu2017video,xue2019video,yuan2019zoom,park2020bmbc} as a guidance of pixel-level motion interpolation. However, integrating a pre-trained flow model makes the whole architecture cumbersome, while with only pixel-level information the task-oriented flow alone is still insufficient in handling complex occlusion and blur. As opposed to this, kernel-based methods~\cite{niklaus2017video,niklaus2017video_sepcov,peleg2019net} synthesize the intermediate frames by convolution operations over local patches surrounding each output pixel. Nevertheless, it cannot deal with large motions beyond the kernel size and it typically suffers from high computational cost. There are also hybrid methods~\cite{bao2019depth,bao2019memc} that combine the advantages of flow-based and kernel-based methods, but the 
networks are much heavier and thus limit their applications.

We observe a growing tendency that more and more complicated and heavy DNN-based models are designed for interpolating video frames. Most of the methods proposed in the past few years~\cite{bao2019depth,bao2019memc,cheng2020video,choi2020channel,jiang2018super,lee2020adacof,niklaus2017video_sepcov,xu2019quadratic} involve training and inference on DNN models consisting of over 20 million parameters. 
For example, the hybrid MEMC-Net~\cite{bao2019memc} consists of more than 70 million parameters and requires around 280 megabytes if stored in 32-bit floating point. Normally, large models are difficult to train and inefficient during inference. Moreover, they are not likely to be deployed on mobile devices, which restricts their scenarios to a great extent. In the mean time, other work~\cite{chi2020all,liu2017video,xue2019video,yuan2019zoom} directly focus on simple and light-weight video interpolation algorithms. However, they either perform less competitively on benchmark datasets or are bound to specific design that lack of transferability.

In this paper, we propose a \emph{compression-driven} network design for video interpolation (CDFI) that takes advantage of model compression~\cite{bucilu2006model,cheng2017survey,zhu2017prune}. To the best of our knowledge, we are the first to explore the \emph{over-parameterization} issue appearing in the state-of-the-art DNN models for video interpolation. Concretely, we compress the recently proposed \mbox{AdaCoF}~\cite{lee2020adacof} via fine-grained pruning~\cite{zhu2017prune} based on sparsity-inducing optimization~\cite{chen2020orthant}, and show that a 10$\times$ compressed \mbox{AdaCoF} is still able to maintain a similar benchmark performance as before, indicating a considerable amount of redundancy in the original model. \emph{The compression provides us two direct benefits: (i) it helps us understand the model architecture in depth, which in turn inspires an efficient design; (ii) the obtained compact model makes more room for further improvements that could potentially boost the performance to a new level.} Towards justifying the latter point, observing that \mbox{AdaCoF} is capable of handling large motion while is short of dealing with occlusion or preserving finer details, we improve upon the compact model by introducing a multi-resolution warping module that utilizes a feature pyramid representation of the input frames
to help with the image synthesis. As a result, our final model outperforms \mbox{AdaCoF} on three benchmark datasets with a large margin (more than 1 dB of PSNR on the Middlebury~\cite{baker2011database} dataset) while is only a quarter of its initial size. Note that typically it is difficult to implement the same improvements on the original heavy model. Experiments show that our model also performs favorably against other state-of-the-art methods. 

\begin{figure}[]
    \centering
    \includegraphics[width=0.45\textwidth]{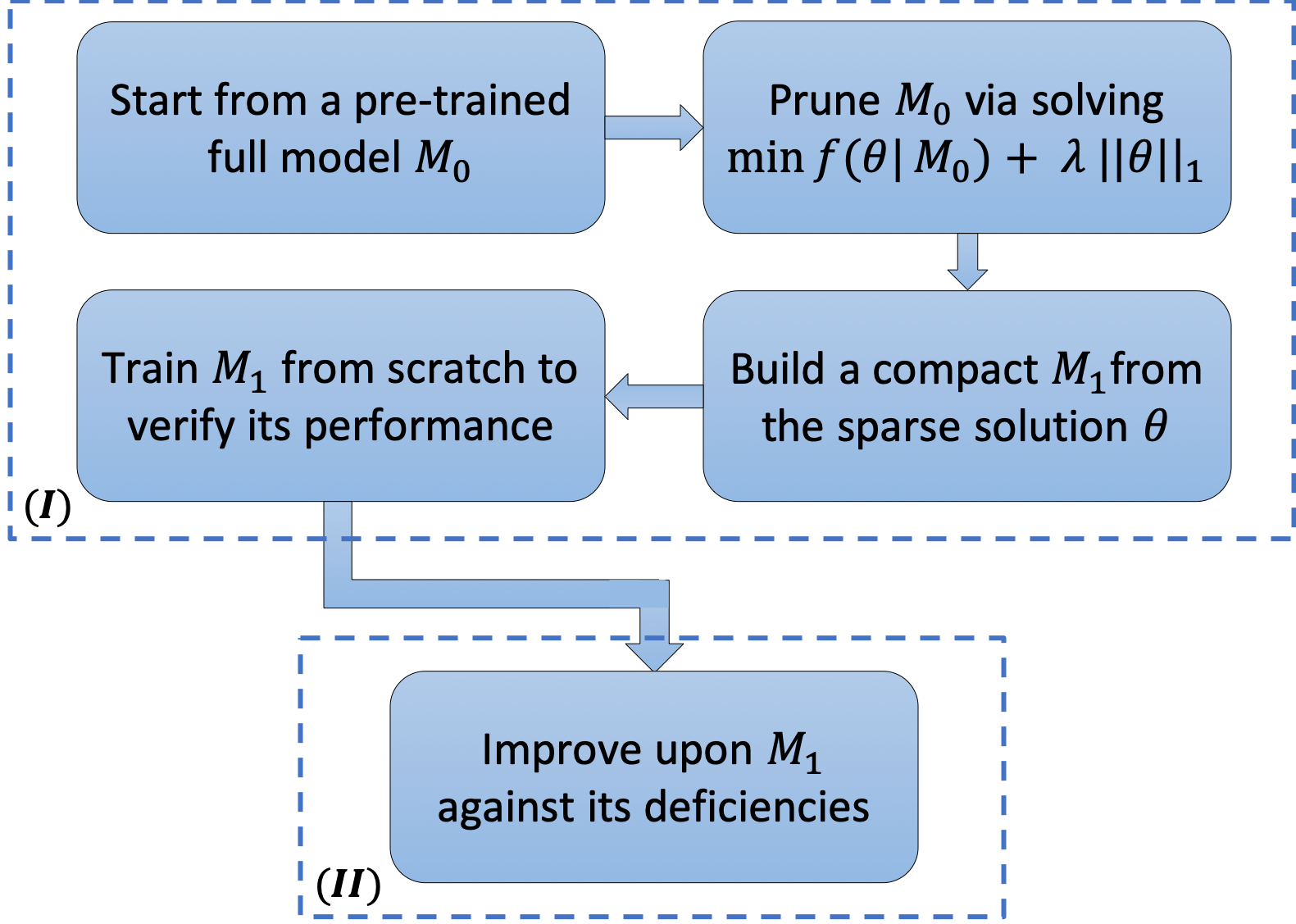}\hfill
     \caption{\textbf{Pipeline of CDFI.} Stage (I): compression of the baseline; Stage (II): improvements upon the compression.}
    \label{fig:compress}
    \vspace{-.15in}
\end{figure}

In short, we present a {compression-driven} framework for video interpolation, in which we take a step back with reflections on over-parameterization. We first compress \mbox{AdaCoF} and obtain a compact model but performs similarly well, then we  improve on top of it. The pipeline of CDFI is illustrated in Figure~\ref{fig:compress}. This retrospective approach leads to superior performance and can be easily transferred to any other DNN-based frame interpolation algorithm.

\section{Related work}

\subsection{Video frame interpolation}


Conventional video frame interpolation is modeled as an image sequence problem, \eg, the path-based~\cite{mahajan2009moving} and phase-based approach~\cite{meyer2018phasenet,meyer2015phase}. Unfortunately, these methods are less effective in complex scenes due to their incapability of accurately estimating the path~(flow) or representing high-frequency components.

Convolutional neural network (CNN) has recently demonstrated its success in understanding temporal motion~\cite{dosovitskiy2015flownet,ilg2017flownet,raket2012motion,sun2018pwc,weinzaepfel2013deepflow,werlberger2011optical} through predicting optical flow, leading to flow-based motion interpolation algorithms. \cite{long2016learning}~trains a deep CNN to directly synthesize the intermediate frame. \cite{liu2017video} estimates the flow by sampling from the 3D spatio-temporal neighborhood of each output pixel. \cite{jiang2018super,park2020bmbc,xue2019video,yuan2019zoom}~utilize bi-directional flows to warp frames and resort to additional modules to handle occlusion. \cite{niklaus2018context,niklaus2020softmax} integrate an off-the-shelf flow model~\cite{sun2018pwc} into the network.
Also, ~quadratic~\cite{liu2020enhanced,xu2019quadratic} and cubic~\cite{chi2020all} non-liner models are proposed to approximate complex motions.

One major drawback of the flow-based methods is that only pixel-wise information is used for interpolation. 
In contrast, kernel-based methods propose to generate the image by convolving over local patches near each output pixel. For example, \cite{niklaus2017video_sepcov}~estimates  spatially-adaptive 2D convolution kernels and \cite{niklaus2017video}~improves its efficiency by using pairs of 1D kernels for all output pixels simultaneously.
\cite{bao2019depth,bao2019memc}~integrate both optical flow and local kernels; specifically~\cite{bao2019depth} detects the occlusion with depth information. However, those methods only rely on local kernels and cannot deal with large motion beyond the rectangular kernel region.


Inspired by the flexible spatial sampling locations of deformable convolution (DConv)~\cite{dai2017deformable,zhu2019deformable}, \cite{lee2020adacof}~proposes the AdaCoF model that utilizes a spatially-adaptive separable DConv to synthesize each output pixel. \cite{shi2020video}~generalizes it by allowing sampling in the full spatial-temporal space. \cite{cheng2020video}~is similar to AdaCoF except that it estimates 1D separable kernels to approximate 2D kernels. \cite{cheng2020multiple}~extends~\cite{cheng2020video} to produce the intermediate frame at arbitrary time step. This paper is also based on AdaCoF; however, unlike the previous work, for the first time we explore the over-parameterization issue presenting in the existing DNN-based approaches, and show that a much smaller model performs similarly well through compression. Moreover, by addressing its drawbacks upon the compression, one can easily build a model (still small) so that it outperforms the original one to a large extent. This compression-driven network design  is generic and can be transferred to any other DNN-based frame interpolation algorithms.

\subsection{Pruning-based model compression}

Model compression~\cite{bucilu2006model,cheng2017survey} is particularly important to DNN models, which are known to suffer high cost of storage and computation. In general, model compression can be categorized into several types: pruning~\cite{zhu2017prune}, quantization~\cite{polino2018model}, knowledge distillation~\cite{hinton2015distilling} and AutoML~\cite{he2018amc}. We adopt the pruning technique for its simplicity, which seeks to induce sparse connections. There are many hybrid pruning methods~\cite{chen2015compressing,han2015deep,ullrich2017soft} that are suitable for model deployment,
but they may be overkill for our purpose of searching and designing the architecture \emph{after the compression}. That being said,  compression plays a completely different role in our work, namely it works as a tool for a better understanding of the underlying architecture and makes room for further improvements. For this reason, we turn our attention to optimization-based sparsity-inducing pruning techniques~\cite{lebedev2016fast,li2016pruning,wen2016learning,zhou2016less} which involve training with sparsity constraints, \eg $\ell_0$ or $\ell_1$ regularizers. Specifically, we use a simple three-step pipeline (see Stage (I) in Figure~\ref{fig:compress}) which is most similar to~\cite{chen2020neural,han2015learning} that involves: (i) training with $\ell_1$-norm sparsity constraint; (ii) reformulating a small dense network according to the sparse structures identified in each layer; and (iii) retraining the small network to verify its performance. We will see shortly (Sec. \ref{sec:compression}) that its implementation and test is straightforward.

\begin{figure}[]
{\small \hspace{-.2in} Ground-truth\hspace{.2in} AdaCoF~\cite{lee2020adacof}\hspace{.3in} Ours}
    \centering
    \includegraphics[width=0.36\textwidth]{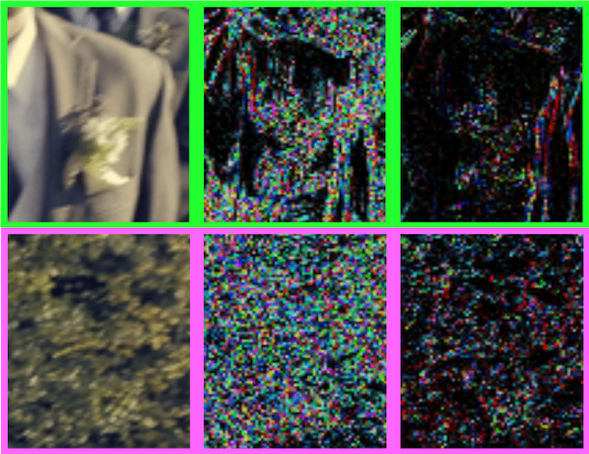}
    \caption{\textbf{Visualization of the difference between the interpolation and the ground-truth image.}}
    \label{fig:feat}
    \vspace{-.15in}
\end{figure}

\begin{figure*}[!ht]
\vspace{-.1in}
    \centering
    \includegraphics[width=0.99\textwidth]{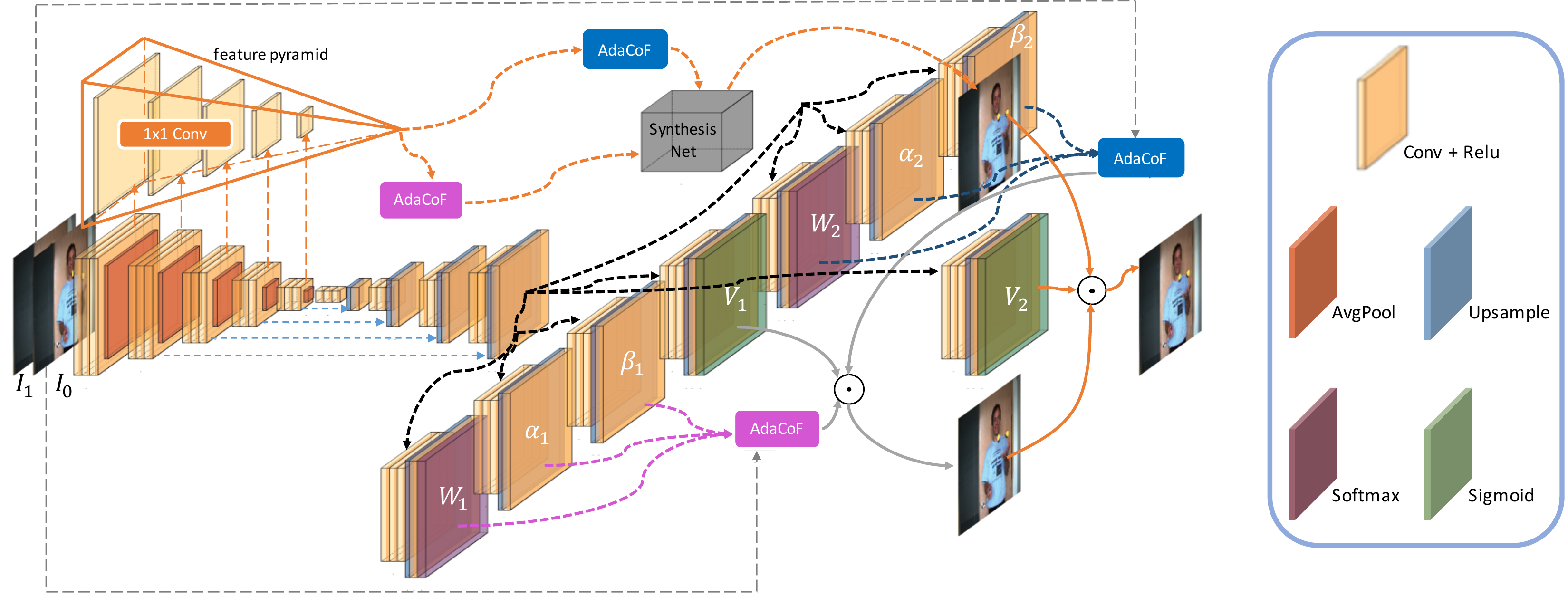}
    \caption{\textbf{Illustration of our architecture design based on the compressed AdaCoF~\cite{lee2020adacof}.} The lower part (AdaCoF) consists of a U-Net, a group of sub-networks for estimating two sets of $\{W_i,\alpha_i,\beta_i\}$ in AdaCoF operation~\eqref{eq:adacof} correspond to backward/forward warping, and an occlusion mask $V_1$ for synthesizing one candidate intermediate frame $I_{0.5}^{(1)}$. The upper part (our design) extracts a feature pyramid representation of the input frames through 1-by-1 convolutions from the encoder of the U-Net, then the multi-scale features are warped by AdaCoF operation of learned backward/forward parameters, which are fed into a synthesis network to generate another candidate intermediate frame $I_{0.5}^{(2)}$. Note that the pink and blue AdaCoF modules are associated with $\{W_1,\alpha_1,\beta_1\}$ and $\{W_2,\alpha_2,\beta_2\}$, respectively. Finally, the network outputs the interpolation frame by blending $I_{0.5}^{(1)}$ and $I_{0.5}^{(2)}$ via an extra occlusion mask $V_2$.}
    \label{fig:model}
    \vspace{-.15in}
\end{figure*}

\section{The proposed approach}

Given two consecutive frames $I_0$ and $I_1$ in a video sequence, the goal of video frame interpolation is to synthesize an intermediate frame $I_t$, where $t\in(0,1)$ is an arbitrary temporal position. 
A common practice is $t = 0.5$, that is synthesizing the middle frame between $I_0$ and $I_1$. We now introduce the proposed CDFI framework with the recently proposed AdaCoF~\cite{lee2020adacof} as an instance.

\subsection{Motivation}
 
To describe AdaCoF, we begin with the introduction of one of its key components, a spatially-adaptive separable DConv operation for synthesizing one image (denoted by $I_{\text{out}}$) from another one (denoted by $I_{\text{in}}$). 
Towards synthesizing  $I_{\text{out}}$ from $I_{\text{in}}$,  the input image $I_{\text{in}}$ is padded such that $I_{\text{out}}$ preserves the original shape of $I_{\text{in}}$.  For each pixel $(i,j)$ in $I_{\text{out}}$, AdaCoF computes $I_{\text{out}}(i, j)$  by convolving a deformable patch surrounding the reference pixel $(i,j)$ in $I_{\text{in}}$:
 
 \vspace{-.2in}
\begin{align}\label{eq:adacof}
\sum_{k=0}^{F-1}\sum_{l=0}^{F-1}
W_{i,j}^{(k,l)}I_{\text{in}}\big(i+dk+\alpha_{i,j}^{(k,l)},j+dl+\beta_{i,j}^{(k,l)}\big),
\end{align}
\vspace{-.1in}

\noindent 
where $F$ is the deformable kernel size, $W_{i,j}^{(k,l)}$ is the $(k,l)$-th kernel weight in synthesizing $I_{\text{out}}(i, j)$, $\vec\Delta:=\big(\alpha_{i,j}^{(k,l)},\beta_{i,j}^{(k,l)}\big)$ is the offset vector of the $(k,l)$-th sampling point associated with $I_{\text{in}}(i, j)$, and $d\in\{0,1,2,\cdots\}$ is the dilation parameter that helps to explore a wider area. Note that the values of $F$ and $d$ are pre-determined. For synthesizing each output pixel in $I_{\text{out}}$, a total number of $F^2$ points are sampled in $I_{\text{in}}$. With the offset vector $\vec\Delta$, the $F^2$ sampling points are not necessarily restricted inside a rigid rectangular region centered at the reference point. On the other hand, unlike the classic DConv, AdaCoF uses different kernel weights across different reference pixels $(i,j)$, indicated by $W_{i,j}^{(k,l)}$ in~\eqref{eq:adacof}; hence the attribute ``separable''~\cite{niklaus2017video_sepcov}.



Although AdaCoF is flexible in handling large and complex motion since the parameters $\{W_{i,j}^{(k,l)}, \alpha_{i,j}^{(k,l)},\beta_{i,j}^{(k,l)}\}$ are computed specifically for each pair of input frames, it is unable to deal with severe occlusion and non-stationary finer details, as shown in Figure~\ref{fig:example}. We further visualize the difference between the interpolation and the ground-truth in Figure~\ref{fig:feat}. AdaCoF is insufficient in preserving the contextual information because the interpolation is simply obtained by blending the two warped frames through a sigmoid mask ($V_1$), as demonstrated in Figure~\ref{fig:model}. A natural question to ask is that if we can make direct improvements on top of it. However, we find the architecture design of the AdaCoF model is relatively cumbersome, especially the encoder-decoder part. For example, six $512\times512\times 3\times 3$ convolutional layers are employed in the middle, which is an entire heuristic since it is unclear whether this design is sufficient or not for the interpolation task. The original AdaCoF model has 21.8 millions of parameters when $F=5,d=1$  and requires 83.4 megabytes if stored with PyTorch. Typically, such a large model takes a long time for training and validation, and thus prohibits direct improvements upon it. Towards better understanding the architecture and improving its performance, we propose a compression-driven approach described as follows.

\subsection{First stage: compression of the baseline}\label{sec:compression}

As the first stage in our approach, we compress the baseline model, \ie, AdaCoF here, by leveraging the fine-grained model pruning~\cite{zhu2017prune} through sparsity-inducing optimization~\cite{chen2018fast}. Specifically, given a pre-trained full model $M_0$, we start by re-training (fine-tuning) its weights~$\theta$ by imposing an $\ell_1$ norm sparsity regularizer, and solve the following optimization problem:
\begin{align}\label{eq:sparse_prob}
 \min_{\theta}\ f(\theta | M_0) + \lambda \|\theta\|_1,
\end{align}
where $f(\cdot)$ denotes the training objective for our task (see Sec.~\ref{sec:training} for details) and $\lambda>0$ is the regularization constant. It is known that with appropriately chosen $\lambda$ the formulation~\eqref{eq:sparse_prob} promotes a sparse solution, with which one can easily identify those important connections among neurons, namely the ones corresponding to non-zero weights. Towards solving~\eqref{eq:sparse_prob}, we utilize the newly proposed orthant-based stochastic method~\cite{chen2020orthant} for its efficient mechanism in promoting sparsity and less performance regression compared with other solvers. By solving the $\ell_1$-regularized problem~\eqref{eq:sparse_prob}, we indeed perform a fine-grained pruning since zeros are promoted in an unstructured manner. Note that one can also impose group sparsity constraints~\cite{chen2020half,lebedev2016fast,zhou2016less}, \eg, mixed $\ell_1/\ell_2$, to prune the kernel weights in a group-wise fashion. We only adopt the $\ell_1$ 
constraint in the presentation for its simplicity.

After obtaining a sparse solution $\hat \theta$, different than~\cite{han2015learning} that directly operates on the sparse network, we re-design a small dense network $M_1$ based on the sparsity computed in each layer. Given the $l$-th convolutional layer consisting of $K_l=C_l^{\text{in}}\times C_l^{\text{out}}\times q\times q$ parameters (denoted as $\hat\theta_l$), where $C_l^{\text{in}}$ is the number of input channels, $C_l^{\text{out}}$ is the number of output channels, $q\times q$ is the kernel size, then the sparsity $s_l$ and density ratio $d_l$ of this layer are respectively defined as 

\vspace{-0.2in}
\begin{align}
 s_l:= \big(\text{\# of zeros in } \hat\theta_l\big)/{K_l}\quad\text{and}\quad
 d_l:= 1-s_l.
\end{align}

\vspace{-0.05in}
\noindent Inspired by~\cite{chen2020neural}, we use $d_l$ as the \emph{compression ratio} and compute $\widetilde C_l^{\text{in}}:=\lceil d_l\cdot C_l^{\text{in}}\rceil $ as the number of kernels we actually need in that layer. The main intuition is that the density ratio $d_l$ reflects the minimal amount of necessary information that needs to be encoded in that layer without affecting performance largely. Since $C_l^{\text{in}}\equiv C_{l-1}^{\text{out}}$, we also update $\widetilde C_{l-1}^{\text{out}}=\widetilde C_l^{\text{in}}$, then repeat the above process for computing the number of kernels in the $(l-1)$-th layer by $\widetilde C_{l-1}^{\text{in}}:=\lceil d_{l-1}\cdot C_{l-1}^{\text{in}}\rceil $, and so on. In words, we reformulate a small network by updating the number of kernels in each convolutional layer according to its density ratio, and proceed from back to the front. Since AdaCoF is fully convolutional (see Figure~\ref{fig:model}), the above procedure can be easily implemented by reducing the number of input/output channels for each layer, leading to a much more compact architecture. In fact, the strategy does not bind to convolutional layers. One can also operate on a fully connected layer by re-computing its number of input/output features accordingly, making it extensible to other architectures.

\begin{figure}[]
\vspace{-.13in}
    \centering
    \includegraphics[width=0.36\textwidth]{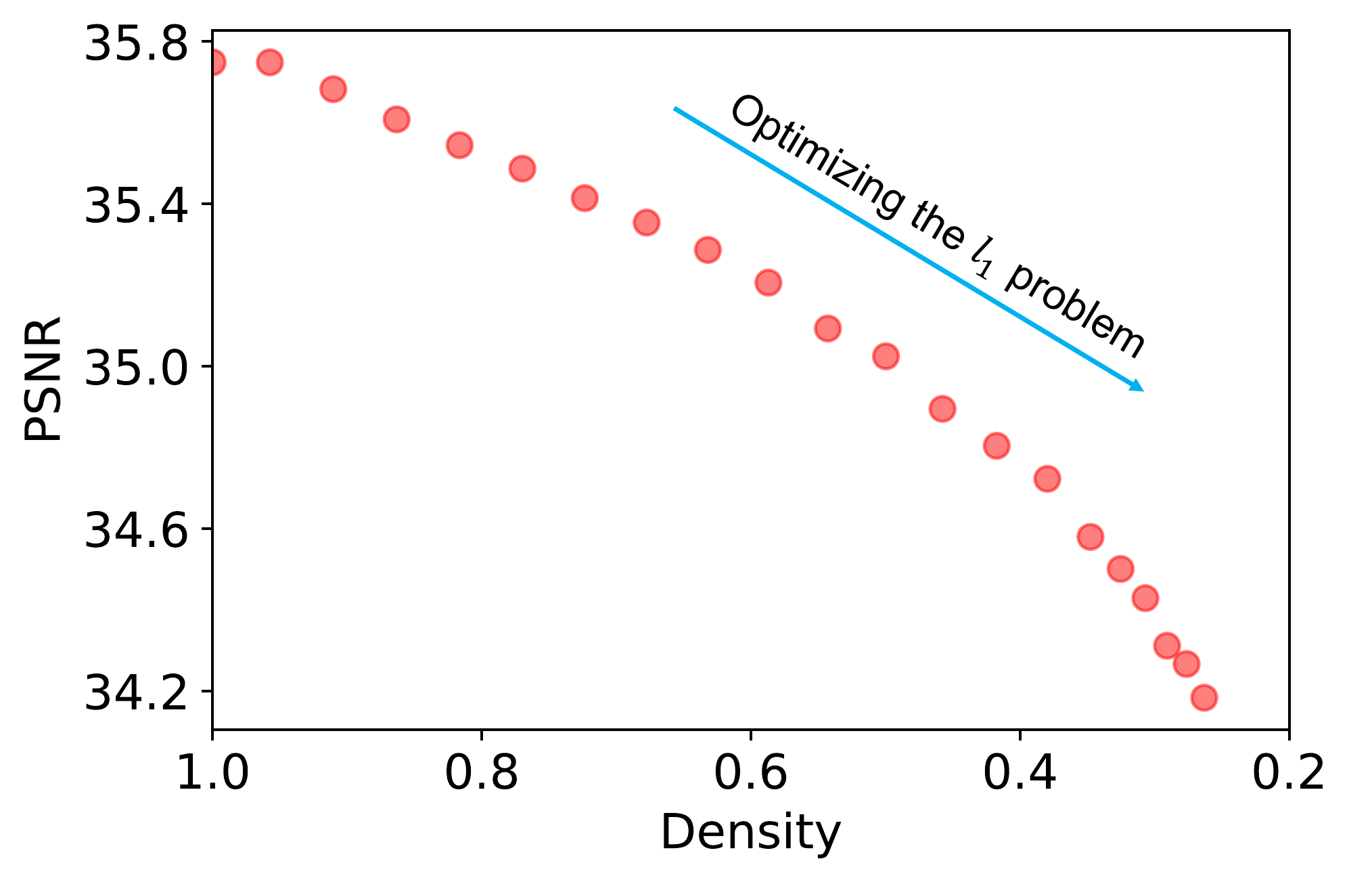}\hfill
    \caption{\textbf{Plot of PSNR against the density of AdaCoF, trained on Middlebury, within 20 epochs of optimizing equation \eqref{eq:sparse_prob}.}}
    \label{fig:prune}
    \vspace{-.2in}
\end{figure}

Finally, we train the compressed model $M_1$ from scratch (without the $\ell_1$ constraint) to verify its performance. Typically, it takes a significantly shorter time than that of the full model $M_0$ due to its compactness. The entire compression pipeline is illustrated as the Stage (I) of Figure~\ref{fig:compress}. We remark that a pre-trained $M_0$ is not necessarily required for the sake of compression since problem~\eqref{eq:sparse_prob} is suitable for a one-shot training/pruning, but $M_0$ allows us to make sure the compressed model works competitively as before.

\myparagraph{Compression of AdaCoF.} We now apply the compression strategy to the AdaCoF model~\cite{lee2020adacof}. We use the pre-trained model provided by the authors. Starting with the $\ell_1$-regularized problem~\eqref{eq:sparse_prob}, where $\lambda$ is set as $10^{-4}$, we run the orthant-based stochastic solver~\cite{chen2020orthant} for 20 epochs by feeding the model with only 1000 video triplets from Vimeo-90K~\cite{xue2019video}. For each epoch, we record the network density and the PSNR evaluated on the Middlebury dataset~\cite{baker2011database}, as plotted in Figure~\ref{fig:prune}. One can see that the model performance declines as more sparsity is promoted. After 20 epochs of training, the density of the network is down to 26\%. Interestingly, by examining the density ratio of each layer, we find that the six $512\times 512\times 3\times 3$ convolutional layers in the middle of the U-Net are among the most redundant portions. In particular, the $512\times 512\times 3\times 3$ convolutional layer following the upsampling layer achieves a density ratio of only 7\%, which means 93\% of the kernel is of little use. This observation confirms our previous conjecture that the original architecture design has a considerable amount of redundancy. Then we reformulate a compact network guided by the computed density ratio in each layer, as described before, and train it from scratch by using the entire taining set (51312 video triplets) of Vimeo-90K. In this case, the training of the compressed AdaCoF is about 5$\times$ faster than previously. When it finishes, we compare the before-and-after models in Table~\ref{tab:compress}, where PSNR and SSIM~\cite{wang2004image} are evaluated on the Middlebury dataset, and time and FLOPS are calculated in synthesizing a $3\times 1280\times 720$ frame on RTX 6000 Ti GPU. Note that although the PSNR drops below 34.2 during the $\ell_1$ optimization, after formulating and training the compressed model rises again to 35.46, which is on par with the original uncompressed AdaCoF. We conclude that a 10$\times$ compressed AdaCoF still maintains a similar performance as its original counterpart.

\begin{table}[]\small
\vspace{-.1in}
\begin{center}
\begin{tabular}{ccc}
\toprule               & \begin{tabular}[c]{@{}c@{}}Original AdaCoF\\ ($F=5,d=1$)\end{tabular} & \begin{tabular}[c]{@{}c@{}}After\\ Compression\end{tabular} \\ \midrule
PSNR                                                           & \color{red}\textbf{35.72}       & 35.43             \\
SSIM                                                           & \color{red}\textbf{0.96}        & \color{red}\textbf{0.96}             \\
Size (MB)                                                      & 83.4        & \color{red}\textbf{9.4}               \\
Time (ms)                                                      & 82.6        & \color{red}\textbf{60.4}              \\
FLOPS (G)                                                      & 359.2       & \color{red}\textbf{185.9}             \\
 \begin{tabular}[c]{@{}c@{}}Parameters (M)\end{tabular} & 21.8        & \color{red}\textbf{2.45}              \\
 \bottomrule
\end{tabular}
\end{center}
\vspace{-.1in}
\caption{\textbf{
The statistics of AdaCoF and the compressed version.
}}
\label{tab:compress}
\vspace{-.2in}
\end{table}

\subsection{Second stage: improve upon the compression}\label{sec:improve}

In the second stage, we improve upon the compression against its deficiencies. The point is that the compression makes room for further improvements due to its compactness, which is typically difficult if directly operating on the original large model, \eg, the long training and validation time appears daunting in the first place.

Observing that AdaCoF is short of handling severe occlusion and preserving finer details, we design three specific components, \ie, a feature pyramid, an image synthesis network and a path selection mechanism, on top of the compressed AdaCoF. Note that the improvements are case by case since different baselines have their own weakness. 

\begin{table*}[!ht]\footnotesize
\vspace{-.1in}
\begin{center}
\begin{tabular}{lcccccccccc}
\toprule 
\multirow{2}{*}{}                                                                  & \multicolumn{3}{c}{Vimeo-90K~\cite{xue2019video}} & \multicolumn{3}{c}{Middlebury~\cite{baker2011database}} & \multicolumn{3}{c}{UCF101-DVF~\cite{liu2017video}} & \multirow{2}{*}{\begin{tabular}[c]{@{}c@{}}Parameters\\ (million)\end{tabular}} \\ \cmidrule(lr){2-4} \cmidrule(lr){5-7} \cmidrule(lr){8-10}
                                                                                   & PSNR     & SSIM     & LPIPS    & PSNR      & SSIM     & LPIPS    & PSNR      & SSIM     & LPIPS    &                                                                                 \\ \midrule
AdaCoF ($F=5, d=1$)                                                                  & 34.35    & 0.956    & 0.019    & 35.72     & 0.959    & 0.019    & 35.16     & \color{red}\textbf{0.950}    & 0.019    & 21.84                                                                            \\ 
Compressed AdaCoF ($F=5, d=1$)     &     34.10     &    0.954      &     0.020     &    35.43       &   0.957       &     0.018     &    35.10       &     \color{red}\textbf{0.950}     &   0.019       & \color{red}\textbf{2.45}   \\ \midrule 
AdaCoF+ ($F=11, d=2$)        &     34.56     &    0.959      &     0.018     &     36.09      &   0.962       &     0.017     &     35.16      &     \color{red}\textbf{0.950}     &    0.019      &           22.93                  \\ 
Compressed AdaCoF+ ($F=11, d=2$)    &      34.44    &   0.958       &    0.019      &  35.73         &  0.960        &  0.018        &    35.13       &   \color{red}\textbf{0.950}       &    0.019      &  2.56  \\  \midrule
\begin{tabular}[c]{@{}c@{}}Ours: FP ($F=5, d=1$)\end{tabular} &    34.62      &    0.962      &    0.011      &    36.13       &  0.961        &    0.008      &    35.08       &       0.949   &   \color{red}\textbf{0.015}        &   4.88                                                                              \\
Ours: FP + 1x1 Conv ($F=5, d=1$)    &  34.82        &  0.963        &   0.011       &  36.52         & 0.964         &   0.008       &     35.11      &    0.949      &     \color{red}\textbf{0.015}     &    4.72                                                                             \\ 
Ours: FP + 1x1 Conv ($F=11, d=2$)                              &    34.82      &   0.963       &   0.011       &  36.70         &  0.964        &  0.008        &  35.14         &     0.949     &  \color{red}\textbf{0.015}          & 4.83                                                                                \\ 
Ours: FP + 1x1 Conv + PS ($F=11, d=2$)   &   \color{red}\textbf{35.17}       &    \color{red}\textbf{0.964}      &    \color{red}\textbf{0.010}      &  \color{red}\textbf{37.14}         &  \color{red}\textbf{0.966}        &   \color{red}\textbf{0.007}       &   \color{red}\textbf{35.21}        &  \color{red}\textbf{0.950}        &  \color{red}\textbf{0.015}        &       4.98                                                                          \\ \bottomrule
\end{tabular}
\end{center}
\vspace{-.1in}
\caption{\textbf{Ablation experiments on the architecture design of our approach.}}
\label{tab:ablation}
\vspace{-.05in}
\end{table*}

\begin{figure*}[]
\vspace{-.08in}
    \centering
    \includegraphics[width=0.49\textwidth]{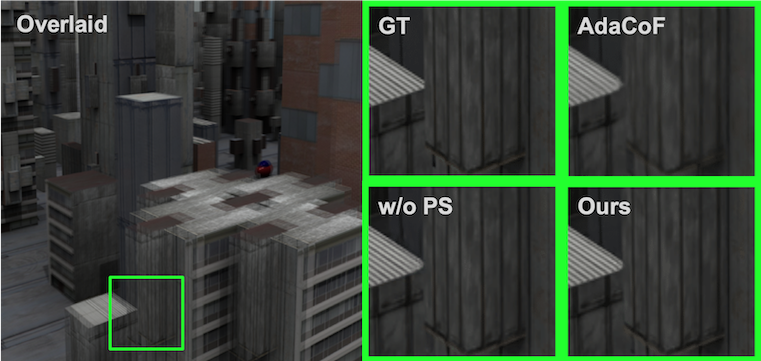} \ \ 
    \includegraphics[width=0.49\textwidth]{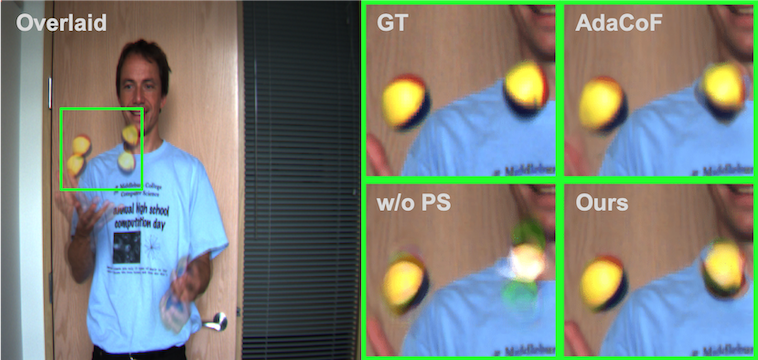}
    \vspace{-.05in}
     \caption{\textbf{Examples of adding the path selection (PS) mechanism in our design.}}
    \label{fig:PS}
    \vspace{-.2in}
\end{figure*}

\myparagraph{Feature pyramid.} In AdaCoF, the final interpolation frame is computed by blending the two warped frames through a single sigmoid mask $V_1$ (see Figure~\ref{fig:model}), which is a generalization of using a binary mask to determine the occlusion weights of the two warped frames for each output pixel. We argue that with only raw pixel information the loss of  contextual details in the input frames is inevitable since it lacks guidance from the feature space. Instead, we extract a feature pyramid representation~\cite{niklaus2020softmax} of the input frames from the encoder part of the U-Net. Specifically, it has five feature levels in accordance with the encoder, and for each level we utilize a 1-by-1 convolution to filter the encoder at multi-scale with $4,8,12,16,20$ output features (in descending order by the feature scale). The extracted multi-scale features are then warped by AdaCoF operation~\eqref{eq:adacof}, which captures the motion in the feature space.

\myparagraph{Image synthesis network.} To better make use of the extracted multi-scale features, we resort to a GridNet~\cite{fourure2017residual} architecture with three rows and six columns in synthesizing the image, which is also employed in~\cite{niklaus2018context,niklaus2020softmax} for its superiority in combining multi-scale information.  Particularly, we feed the synthesis network with both the forward- and backward-warped multi-scale feature maps, generating a single RGB image that focuses on the contextual details.

\myparagraph{Path selection.} In order to take advantage of both AdaCoF (handling complex motion) and our own components (handling contextual details), we apply a path selection mechanism in generating the final interpolation result. As shown in Figure~\ref{fig:model}, one path leads to the output of the original AdaCoF (denoted as $I_{0.5}^{(1)}$), which is computed by blending two warped input frames using the occlusion mask $V_1$. Parallel to this, another path leads to the output of the synthesis network (denoted as $I_{0.5}^{(2)}$), which is computed by combining the warped multi-scale feature maps. In the end, we learn another occlusion module $V_2$ to synthesize the final result from $I_{0.5}^{(1)}$ and $I_{0.5}^{(2)}$, and we expect that $I_{0.5}^{(2)}$ can compensate for the lack of contextual information in $I_{0.5}^{(1)}$.

The above three specific components can not only be easily incorporated into the compressed AdaCoF, but also boost the performance to a large extent while still maintain the compactness (see Sec.~\ref{sec:exp}).

\begin{table*}[!ht]\footnotesize
\vspace{-.15in}
\begin{center}
\begin{tabular}{lccccccccccc}
\toprule 
\multirow{2}{*}{}              & \multirow{2}{*}{\begin{tabular}[c]{@{}c@{}}Training\\ dataset\end{tabular}}                                                     & \multicolumn{3}{c}{Vimeo-90K~\cite{xue2019video}} & \multicolumn{3}{c}{Middlebury~\cite{baker2011database}} & \multicolumn{3}{c}{UCF101-DVF~\cite{liu2017video}} &
\multirow{2}{*}{\begin{tabular}[c]{@{}c@{}}Parameters\\ (million)\end{tabular}} \\ \cmidrule(lr){3-5} \cmidrule(lr){6-8} \cmidrule(lr){9-11}
&  & PSNR      & SSIM    & LPIPS   & PSNR    & SSIM    & LPIPS    & PSNR     & SSIM    & LPIPS    &              \\ 
\midrule
$^\dagger$SepConv - $\mathcal{L}_1$~\cite{niklaus2017video_sepcov}     & proprietary & 33.80    & 0.956    & 0.027    & 35.73     & 0.959    & 0.017    & 34.79     & 0.947    & 0.029    & 21.6                                                                            \\ 
$^\dagger$SepConv - $\mathcal{L}_F$~\cite{niklaus2017video_sepcov}         & proprietary &   33.45       &    0.951      &    0.019      &    35.03       &  0.954        &  0.013        &   34.69        &  0.945        &  0.024        &      21.6                                                                           \\ 
$^\dagger$CtxSyn - $\mathcal{L}_{Lap}$~\cite{niklaus2018context}      & proprietary &    34.39      &     0.961     &     0.024     &     36.93      &    0.964      &    0.016      &    34.62       &   0.949       &   0.031       &     --                                                                            \\ 
$^\dagger$CtxSyn - $\mathcal{L}_{F}$~\cite{niklaus2018context} &   proprietary       &    33.76      &    0.955      &    0.017       &   35.95       &   0.959       &   0.013        &      34.01    &      0.941    &  0.024    &        --             \\
$^\dagger$SoftSplat - $\mathcal{L}_{Lap}$~\cite{niklaus2020softmax}    &   Vimeo-90K         &     \color{red}\textbf{36.10}     &    \color{red}\textbf{0.970}      &    0.021       &   \color{red}\textbf{38.42}       &  \color{red}\textbf{0.971}        &   0.016        &      \color{red}\textbf{35.39}    &      \color{red}\textbf{0.952}    &      0.033             &     --     \\ 
$^\dagger$SoftSplat - $\mathcal{L}_{F}$~\cite{niklaus2020softmax}          &     Vimeo-90K       &   \color{blue}\textbf{35.48}       &    \color{blue}\textbf{0.964}         &    \color{blue}\textbf{0.013}       &   \color{blue}\textbf{37.55}       &  0.965        &   \color{blue}\textbf{0.008}        &  35.10        &  0.948    &  0.022   &  --  \\ 
$^\dagger$DAIN~\cite{bao2019depth}   &    Vimeo-90K        &    34.70      &  \color{blue}\textbf{0.964}           &      0.022     &     36.70     &     0.965     &     0.017      &        35.00  &    \color{blue}\textbf{0.950}     &    0.028  &       24.02       \\ 
AdaCoF~\cite{lee2020adacof}       &     Vimeo-90K       &    34.35    & 0.956    & 0.019    & 35.72     & 0.959    & 0.019    & 35.16     & \color{blue}\textbf{0.950}   & \color{blue}\textbf{0.019}    & 21.84                           \\ 
AdaCoF+~\cite{lee2020adacof}       &     Vimeo-90K       &    34.56     &    0.959      &     0.018     &     36.09      &   0.962       &     0.017     &     35.16      &     \color{blue}\textbf{0.950}     &    \color{blue}\textbf{0.019}     &           22.93             \\
EDSC - $\mathcal{L}_{C}$~\cite{cheng2020multiple}       &     Vimeo-90K       &    34.86      &     0.962     &     0.016      &   36.76     &   \color{blue}\textbf{0.966}  &    0.014    &      35.17    &  \color{blue}\textbf{0.950}          & \color{blue}\textbf{0.019}             &   8.9   \\ 
EDSC - $\mathcal{L}_{F}$~\cite{cheng2020multiple}        &     Vimeo-90K       &    34.57      &    0.958      &    \color{red}\textbf{0.010}       &   36.48       &      0.963    &     \color{red}\textbf{0.007}      &   35.04       &  0.948  &  \color{red}\textbf{0.015}       &   8.9    \\
BMBC~\cite{park2020bmbc}       &         Vimeo-90K             &   35.06       &    \color{blue}\textbf{0.964}      &    0.015      &  36.79       &     0.965    &    0.015      &   35.16        &      \color{blue}\textbf{0.950}    &   \color{blue}\textbf{0.019}       &    11.0  \\
CAIN~\cite{choi2020channel}       &         Vimeo-90K             &   34.65       &     0.959     &     0.020     &   35.11   &    0.951      &   0.019       &      34.98     &     \color{blue}\textbf{0.950}     &     0.021     & 42.8  \\
Ours           &        Vimeo-90K                      &   {35.17}       &   \color{blue}\textbf{0.964}     &    \color{red}\textbf{0.010}      &  {37.14}         &   \color{blue}\textbf{0.966}        &   \color{red}\textbf{0.007}       &   \color{blue}\textbf{35.21}        &  \color{blue}\textbf{0.950}       &  \color{red}\textbf{0.015}        &       \color{red}\textbf{4.98}         \\
\bottomrule
\end{tabular}
\end{center}

\vspace{-.12in}
\caption{\textbf{Quantitative comparisons with state-of-the-art methods.} The results of methods marked with $^\dagger$ are cloned from~\cite{niklaus2020softmax}.}
\label{tab:quant}
\vspace{-.2in}
\end{table*}

\subsection{Training} \label{sec:training}

With the architecture described above, we train it using AdaMax~\cite{kingma2014adam} with $\beta_1=0.9, \beta_2=0.999$, an initial learning rate of 0.001 which decays half every 20 epochs, a mini-batch size of 8, and a maximum training epochs of 100.

\myparagraph{Objective function.} Given the interpolated frame $I_{\text{out}}$ of our network and its ground truth $I_{\text{gt}}$, we first employ the Charbonnier penalty~\cite{liu2017video} as a surrogate for the $\ell_1$ loss:
\begin{align}
    \mathcal{L}_{\text{Charbon}}=\rho(I_{\text{out}}-I_{\text{gt}})
\end{align}
where $\rho(\bm x)=(\|\bm x\|_2^2+\epsilon^2)^{1/2}$ and $\epsilon$ is set to $0.001$. Next, we follow~\cite{lee2020adacof} and use a perceptual loss with feature $\phi$ extracted from  \texttt{conv4\_3} of the pre-trained VGG16~\cite{simonyan2014very}:
\begin{align}
 \mathcal{L}_{\text{vgg}}=\|\phi(I_{\text{out}})-\phi(I_{\text{gt}})\|_2.
\end{align}
Then, inspired by the implementation of AdaCoF, we use a total variation loss imposed on the offset vectors for ensuring spatial continuity and smoothness:
\begin{align}
\mathcal{L}_{\text{tv}} = \tau(\bm\alpha_1)+\tau(\bm\alpha_2)+\tau(\bm\beta_1)+\tau(\bm\beta_2)
\end{align}
where $\tau(I)=\sum_{i,j}\rho(I_{i,j+1}-I_{i,j})+\rho(I_{i+1,j}-I_{i,j})$, and $\bm\alpha_1,\bm\alpha_2,\bm\beta_1,\bm\beta_2$ are the offsets modules computed within our network. Lastly, we formulate our final loss function as
\begin{align}
\mathcal{L}= \mathcal{L}_{\text{Charbon}}+ \lambda_{\text{vgg}}\mathcal{L}_{\text{vgg}} + \lambda_{\text{tv}}\mathcal{L}_{\text{tv}}
\end{align}
where we set $\lambda_{\text{vgg}}=0.005,\lambda_{\text{tv}}=0.01$ in the experiments.

\myparagraph{Training dataset.} We use the Vimeo-90K dataset~\cite{xue2019video} for training, which contains 51312/3782 video triplets of size 256$\times 448$ for training/validation. We further augment the data by randomly flipping them horizontally and vertically as well as perturbing the temporal order.

\myparagraph{Evaluation.} Besides the validation set of Vimeo-90K, we also evaluate the model on the well-known Middlebury dataset~\cite{baker2011database} and UCF101~\cite{liu2017video,soomro2012ucf101}. The metrics we use are PSNR, SSIM~\cite{wang2004image} and LPIPS~\cite{zhang2018unreasonable}. Note higher values of PSNR and SSIM indicate better performance, while for LPIPS a lower value corresponds to a better result.

\section{Experiments} \label{sec:exp}

\subsection{Ablation study}

We analyze three components in our proposed method: model compression, feature pyramid, and path selection. 

\myparagraph{Model compression.} As described in Sec.~\ref{sec:compression}, we compress the baseline model to remove a large mount of redundancy, which also facilitates the training and inference. In Table~\ref{tab:ablation}, we compare the performance of the AdaCoF and the compressed counterpart. It shows that a 10$\times$ compressed model does not sacrifice much when evaluated on the three benchmark datasets in different settings of $(F,d)$ (which are the parameters in \eqref{eq:adacof}), revealing the redundancy in AdaCoF and the necessity of the compression stage.

\myparagraph{Feature pyramid.} In order to better capture the contextual details, we incorporate the feature pyramid (FP) module into the compressed AdaCoF followed by warping operations and an image synthesis network (see Sec.~\ref{sec:improve}). We isolate its effect by training a network that only outputs the synthesized image without a path selection mechanism. It turns out that by using only FP module (see ``Ours - FP'', Table~\ref{tab:ablation}) on top of the compressed AdaCoF ($F=5,d=1$), we achieve visible improvements in terms of PSNR, SSIM and LPIPS on the Vimeo-90K and Middlebury datasets. Note that it substantially improves LPIPS on all the three benchmark datasets. Moreover, filtering the multi-scale feature maps with 1-by-1 convolutions leads to better PSNR and SSIM as well as a slightly smaller model size.

\begin{figure*}[!ht]
\vspace{-.2in}
{\footnotesize \hspace{0.85in}Ground-truth\hspace{0.9in} Overlaid \hspace{0.12in}AdaCoF+~\cite{lee2020adacof}\hspace{0.04in}BMBC~\cite{park2020bmbc}\hspace{0.1in}CAIN~\cite{choi2020channel}\hspace{0.07in}EDSC-$\mathcal{L}_C$~\cite{cheng2020multiple}\hspace{0.0in} EDSC-$\mathcal{L}_F$~\cite{cheng2020multiple}\hspace{0.1in} \textbf{Ours}  } \\
    \centering
    \includegraphics[width=0.99\textwidth]{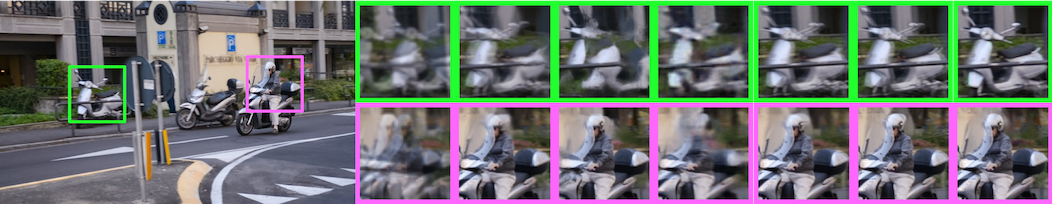}\\
    \includegraphics[width=0.99\textwidth]{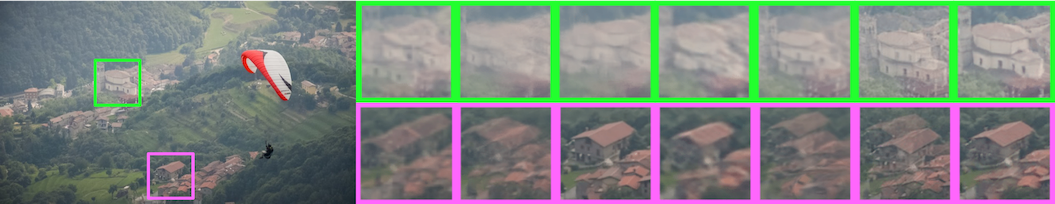}\\
    \includegraphics[width=0.99\textwidth]{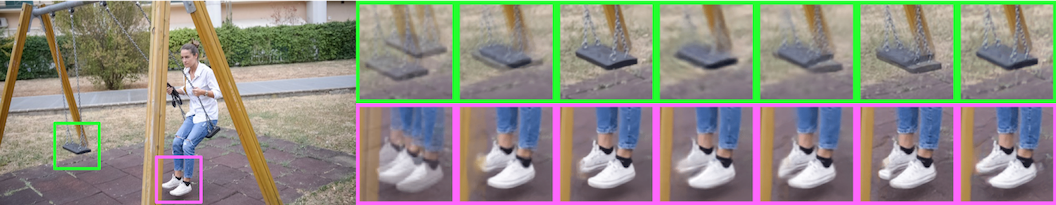}
    \caption{\textbf{Visual comparisons on the DAVIS 2016 dataset~\cite{perazzi2016}.} Our compression-driven method not only outperforms the baseline model AdaCoF but also is more appealing compared with more recently proposed methods in handling large motion, occlusion and fine details.}
    \label{fig:visual}
    \vspace{-.2in}
\end{figure*}

\myparagraph{Path selection.} Although by adding only the FP module (and \mbox{1-by-1} convolutions) we are able to achieve promising quantitative results as shown in Table~\ref{tab:ablation}, it does not take advantage of the capability of AdaCoF in handling complex motion, which can be integrated into our design with the proposed path selection (PS) mechanism. The left example in Figure~\ref{fig:PS} shows that, when there is only fine detail variations in the input frames, adding PS or not does not quite affect our interpolation performance since FP module is capable of synthesizing details (also note the output of AdaCoF is blurry due to the loss of information). On the other hand, the right example contains large motion of two balls, and  with only FP module the model is difficult in capturing the motion of the right ball precisely. In contrast, our final model with the embedded PS mechanism can deal with the large motion very well (even sharper on the edges of the balls compared to AdaCoF). More importantly, our approach preserves the finger shape (see the bottom-left corner) while AdaCoF totally misses that part of information. In conclusion, our completed model with FP and PS can handle both fine details and large motion, and achieves significant improvements when evaluated quantitatively.

\vspace{-.05in}
\subsection{Quantitative evaluation}
\vspace{-.05in}

We evaluate our compression-driven approach based on AdaCoF with $F=11,d=2$ against the other  state-of-the-art DNN methods in Table~\ref{tab:quant}. Since SepCov~\cite{niklaus2017video_sepcov}, CtxSyn~\cite{niklaus2018context} and SoftSplat~\cite{niklaus2020softmax} are not open source, we directly copy their numerical results as well as DAIN's~\cite{bao2019depth} from~\cite{niklaus2020softmax}. For the rest of the methods, we evaluate their pre-trained models on the three datasets. Note that ``AdaCoF'' corresponds to the setting of  $F=5,d=1$ while ``AdaCoF+'' is associated with $F=11,d=2$.

As shown in Table~\ref{tab:quant},  first note that our approach performs favorably against all the compared methods in terms of SSIM and LPIPS. As for PSNR, the proposed method also outperforms all the other methods with a large margin except for SoftSplat~\cite{niklaus2020softmax}. Moreover, our model is significantly smaller than the other competitors. We remark that in the past there do exist some light-weight frame interpolation models, \eg, DVF~\cite{liu2017video}, ToFlow~\cite{xue2019video} and CyclicGen~\cite{liu2019deep}, but they fail to compete with SepConv~\cite{niklaus2017video_sepcov} or CtxSyn~\cite{niklaus2018context} as reported in~\cite{niklaus2020softmax}. Lastly, we observe that AdaCoF~\cite{lee2020adacof} is only mediocre among those methods, but our final model which is based upon it has a significantly better performance while maintains compactness, indicating the superiority of the proposed CDFI design framework.

\vspace{-.01in}
\subsection{Qualitative evaluation}
\vspace{-.01in}

We demonstrate the visual comparisons on the DAVIS dataset~\cite{perazzi2016} in Figure~\ref{fig:visual}. The first and third example contain complex motion and occlusion, 
while the second example involves many non-stationary finer details. Note that AdaCoF+~\cite{lee2020adacof} generates relatively blurry interpolation frame for all these examples (see the motorbike, house and swing stool). In contrast, our method built upon it predicts sharper and more realistic results due to our newly added FP module and PS mechanism. Furthermore, we compare with BMBC~\cite{park2020bmbc}, CAIN~\cite{choi2020channel} and EDSC~\cite{cheng2020multiple}, which are all newly developed within the year. In particular, similar to AdaCoF, EDSC relies on the deformable separable convolution but estimates an extra mask to help with the image synthesis. However, they are not as appealing as our method on the provided examples. One can see that their interpolations normally contain visible artifacts and are not capable of preserving clear details. Note that BMBC~\cite{park2020bmbc} occasionally synthesizes sharp results but is not as consistent as ours. We conjecture that the additional bilateral cost volume in BMBC benefits the intermediate motion estimations, which can also be incorporated into our design. Recall that the size of our model is the smallest among them, which again confirms the advantage of the CDFI network design.

\vspace{-.01in}
\section{Conclusions}
\vspace{-.01in}

We presented a compression-driven network design for frame interpolation (CDFI) that uses model compression as a guide in determining an efficient architecture and then improves upon it. For the first time, we considered the redundancy in the existing methods. As an instance, we showed that a much smaller AdaCoF model performs similarly as the original one, and with simple modifications it is able to outperform the baseline with a large extent and is also superior against other state-of-the-art methods. We emphasize that the optimization-based compression over a baseline model does not rely on particular design of the baseline. 
 Therefore, we believe that our framework is generic to be applied to other models and provides \emph{a new perspective} on developing efficient frame interpolation algorithms. In future work, it will be of interest to construct a better association between the compression and design stages which iteratively refines the underlying architecture.


\newpage

{\small
\bibliographystyle{ieee_fullname}
\bibliography{egbib}

\nocite{reda2019unsupervised}
\nocite{choi2020scene}
\nocite{son2020aim}
}

\end{document}